\documentclass[]{ceurart}

\usepackage{latexsym}
\usepackage{amssymb}
\usepackage{amsmath}
\usepackage{amsthm}
\usepackage{booktabs}
\usepackage{enumitem}
\usepackage{graphicx}
\usepackage{color}
\usepackage[switch]{lineno}
\usepackage{rotating}
\usepackage{booktabs}
\usepackage{soul}
\usepackage{multirow}
\usepackage{listings}
\begin{document}

\copyrightyear{2022}
\copyrightclause{Copyright for this paper by its authors.
  Use permitted under Creative Commons License Attribution 4.0
  International (CC BY 4.0).}

\conference{Preprint} 
  
\title{XEQ Scale for Evaluating XAI Experience Quality}

\author[1]{Anjana Wijekoon}[%
orcid=0000-0003-3848-3100,
email=a.wijekoon@ucl.ac.uk]
\cormark[1]
\fnmark[1]
\address[1]{University College London, London, England}

\author[2]{Nirmalie Wiratunga}[
orcid=0000-0003-4040-2496]
\author[2]{David Corsar}[
orcid=0000-0001-7059-4594]
\author[2]{Kyle Martin}[
orcid=00000-0003-0941-3111]
\author[2]{Ikechukwu Nkisi-Orji}[
orcid=0000-0001-9734-9978]
\address[2]{Robert Gordon University, Aberdeen, Scotland}

\author[3]{Belen Díaz-Agudo}[%
orcid=0000-0003-2818-027X]
\address[3]{Universidad Complutense de Madrid, Spain}

\author[4]{Derek Bridge}[%
orcid=0000-0002-8720-3876]
\address[4]{University College Cork, Ireland}

\cortext[1]{Corresponding author.}
\fntext[1]{This work was done while affiliated with Robert Gordon University, Aberdeen, Scotland.}

\begin{abstract}
Explainable Artificial Intelligence~(XAI) aims to improve the transparency of autonomous decision-making through explanations. Recent literature has emphasised users' need for holistic ``multi-shot'' explanations and personalised engagement with XAI systems. We refer to this user-centred interaction as an XAI Experience. Despite advances in creating XAI experiences, evaluating them in a user-centred manner has remained challenging. In response, we developed the XAI Experience Quality~(XEQ) Scale. XEQ quantifies the quality of experiences across four dimensions: learning, utility, fulfilment and engagement. These contributions extend the state-of-the-art of XAI evaluation, moving beyond the one-dimensional metrics frequently developed to assess single-shot explanations. This paper presents the XEQ scale development and validation process, including content validation with XAI experts, and discriminant and construct validation through a large-scale pilot study. Our pilot study results offer strong evidence that establishes the XEQ Scale as a comprehensive framework for evaluating user-centred XAI experiences.
\end{abstract}

\begin{keywords}
Explanation Experience Quality\sep 
Multi-shot Explanation\sep
Psychometric Theory\sep 
Scale Development
\end{keywords}

\maketitle

\section{Introduction}

\begin{figure*}
    \centering
    \includegraphics[width=\linewidth]{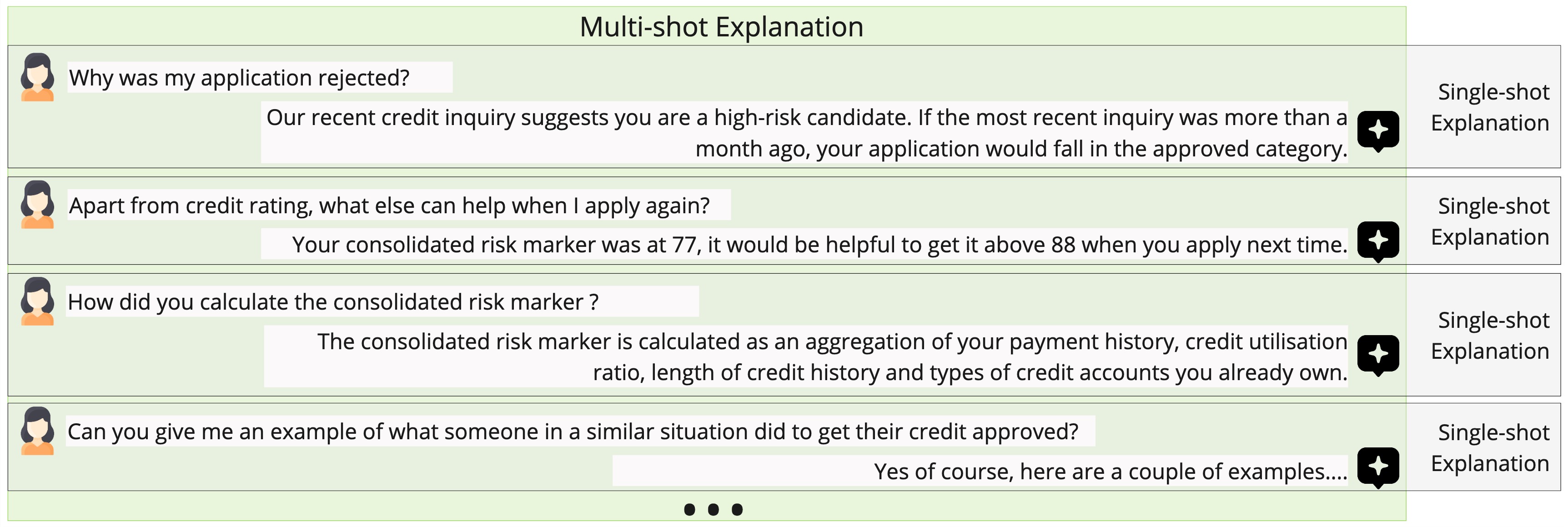}
    \caption{\textcolor{black}{A conversational interaction demonstrating a single-shot explanation (adapted from IBM Explainability 360 platform~\cite{arya2021ai}) extended to a multi-shot XAI experience. Each single-shot explanation is represented by individual turns in the conversation, while a multi-shot explanation is represented by the conversation in its entirety.}}
    \label{fig:multi-single}
\end{figure*}

Explainable Artificial Intelligence (XAI) describes a range of techniques to elucidate autonomous decision-making and the data that informed that AI system~\cite{miller2019explanation,hu2021xaitk,arya2021ai}. Each technique typically provides explanations focusing on a specific aspect of the system and its decisions, often answering a singular question fulfilling a specific intent~\cite{liao2020chi,wijekoon2024tell}. Accordingly, the utility of employing multiple techniques for a holistic explanation of a system becomes increasingly clear~\cite{arya2021ai,wijekoon2023cbr}.
The collection of explanations, provided by different techniques and describing different components of the system, forms what we describe as ``multi-shot'' explanations. 
\textcolor{black}{ Multi-shot explanation treats the elucidation of an AI system as a process which can invoke a variety of knowledge sources to construct explanations that comprehensively answer a user's explanation needs. This is unlike single-shot, which provides a lone explanation that targets a specific aspect of an autonomous decision to resolve an individual query (see Figure~\ref{fig:multi-single}). Furthermore, viewing multi-shot explanations as an interactive process allows users to engage through graphical interfaces~\cite{arya2021ai} or conversations~\cite{malandri2023convxai,wijekoon2023cbr}, enabling personalised experiences tailored to individual needs~\cite{finzel2021explanation}}.

While the utility of user-centred interactive multi-shot explanations is evident in recent literature, evaluating them remains a key research challenge. 
Current works primarily target the development of objective metrics for single-shot techniques~\cite{rosenfeld2021better,yeh2019fidelity}, emphasising the need for reproducible benchmarks on public datasets~\cite{le2023benchmarking}. Such metrics are system-centred and model-agnostic\textcolor{black}{, meaning they are compatible with a variety of XAI systems and} giving the advantage of generalisability. 
However, objective metrics fail to acknowledge the requirements of different stakeholder groups. A satisfactory explanation is reliant on the recipient’s expertise within that domain~\cite{mohseni2021multidisciplinary} and their previous experiences with AI~\cite{kim2023chi,ehsan2024chi}.
Subjective metrics, such as those described in~\cite{hoffman2023measures,holzinger2020measuring}, allow an evaluation which is personalised to the individual and domain. However, existing subjective evaluations lack the capacity to measure the interactive process that underpins multi-shot explanations and how they impact user experience. 
\textcolor{black}{There is a clear need for a tool that facilitates the evaluation of XAI experiences. In turn, this would empower the iterative development of XAI systems to ensure they address the diverse needs of various stakeholder groups.}

\begin{table*}[ht]
\centering
\renewcommand{\arraystretch}{1.15}
\caption{Glossary \textcolor{black}{of term and definitions used throughout this paper.}}
\label{tbl:gloss}
\begin{tabular}{lp{12cm}}
\hline
Term&Definition\\
\hline
XAI System& An automated decision-making system that is designed and developed to provide information about its reasoning.\\
Stakeholder & An individual or group with a vested interest in the XAI system. Stakeholders are a diverse group, encompassing system designers and developers, who hold an interest in the system's technical functionality, the end consumers relying on its decisions, and regulatory authorities responsible for ensuring fair and ethical use.\\
\textcolor{black}{Single-Shot Explanation} & \textcolor{black}{An explanatory process where a lone explanation is provided to answer a specific query.} \\
\textcolor{black}{Multi-Shot Explanation} & \textcolor{black}{An explanatory process where multiple techniques are used to provide two or more alternative and/or complementary insights into a single XAI system or output.} \\
XAI Experience (XE)&A user-centred process of a stakeholder interacting with an XAI system to gain knowledge and/or improve comprehension.\\
XE Quality (XEQ)& The extent to which a stakeholder's explanation needs are satisfied by their XE.\\
\hline
\end{tabular}
\end{table*}

We address this challenge by introducing the XAI Experience Quality~(XEQ) Scale. We define an XAI Experience as the user-centred process of a stakeholder interacting with an XAI system to gain knowledge and/or improve comprehension. XAI Experience Quality (XEQ) is defined as \textit{the extent to which a stakeholder's explanation needs are satisfied by their XAI Experience}. A glossary of all related terminology used throughout this paper is included in Table~\ref{tbl:gloss}. 
Specifically, we ask the research question: ``\textit{How to evaluate an XAI experience, in contrast to assessing single-shot~(non-interactive) explanations?}''. To address this, we follow a formal psychometric scale development process~\cite{boateng2018best} and outline the following objectives: 
\begin{enumerate}
\item conduct a literature review to compile a collection of XAI evaluation questionnaire items;
\item conduct a content validity study with XAI experts to develop the XEQ scale; and 
\item perform a pilot study to refine and validate the XEQ scale for internal consistency, discriminant and construct validity \textcolor{black}{and test-retest reliability.}
\end{enumerate}

The rest of this paper expands on each objective. We discuss related work in Section~\ref{sec:related}. Section~\ref{sec:o1} presents the creation of the initial items bank. The Content Validity study details and results are presented in Section~\ref{sec:o2} followed by Section~\ref{sec:o3} presenting the pilot study for the refinement and validation of the XEQ Scale. Key implications and the practical use of the Scale are discussed in Section~\ref{sec:discuss}. Finally, we offer conclusions in Section~\ref{sec:conc}.

 \section{Related Work}
\label{sec:related}

In the literature, there are several methodologies for developing evaluation metrics or instruments for user-centred XAI.

Hoffman et al.~\cite{hoffman2023measures} employed Psychometric Theory to construct the Satisfaction Scale, evaluating both content validity and discriminant validity. A similar methodology was adopted in~\cite{madsen2000measuring} to develop the Madsen-Gregor human-machine trust scale, relying on pre-existing item lists from previous scales in conjunction with expert insights~\cite{moore1991development}.
Jian et al.~\cite{jian2000foundations} pursued a factor analysis approach involving non-expert users to formulate a human-machine trust scale. They compiled words and phrases associated with trust and its variants, organising them based on their relevance to trust and distrust, which were then clustered to identify underlying factors and formulate corresponding statements. This methodology is particularly suitable in cases where no prior items exist for initial compilation. While these methodologies are robust to produce reliable scales they are resource and knowledge-intensive processes. 

A more frequent approach to scale development is deriving them from existing scales in psychology research. For instance, the System Causability Scale~\cite{holzinger2020measuring} draws inspiration from the widely used System Usability Scale~\cite{brooke1996sus}, while the Hoffman Curiosity Checklist originates from scales designed to assess human curiosity~\cite{hoffman2023measures}. Similarly, the Cahour-Forzy Trust Scale~\cite{cahour2009does} selected questions from research on human trust, and the Hoffman Trust Scale incorporates items from previous trust scales~\cite{cahour2009does,jian2000foundations}. Notably, these derived scales were not evaluated for reliability or other desirable factors, they rely on the quality of the original scales for validity.
In this paper, we opt for the psychometric theory approach to establish the content, construct and discriminant validity of the resulting scale. While this approach is resource-intensive, the complexity and the novelty of the evaluation task necessitate a rigorous approach to scale development.

\section{Initial Items Bank Compilation}
\label{sec:o1}
This section presents the literature review findings that led to the compilation of the initial items bank for the XEQ Scale. 

\subsection{Methodology}
\textcolor{black}{Netemeyer~\cite{netemeyer2003scaling} outlines a framework for developing an item pool, via domain sampling grounded literature review. Following this approach, we conducted a targeted literature review to examine existing research, identify key dimensions, and develop the initial item bank.} 
The reasoning for a targeted review instead of a systematic review is twofold: 1) the purpose of the review is to form the initial item bank which involves in-depth analysis of selected literature~(depth over breadth); and 2) literature under this topic is significantly limited.
The initial findings highlighted that while many publications discuss and emphasise the importance of evaluation dimensions~(what should be or is evaluated), only a few propose and systematically develop metrics for XAI evaluation. 

\subsection{Findings: Evaluation metrics and dimensions}
Hoffman et al.~\cite{hoffman2023measures} are one of the leading contributors and their work has been widely utilised in many user-centred XAI research. They conceptually modelled the ``process of explaining in XAI'' outlining dimensions and metrics for evaluating single-shot explanations from stakeholders' perspectives.
They considered six evaluation dimensions: goodness, satisfaction, mental model, curiosity, trust and performance, For each dimension, they either systematically developed an evaluation metric or critiqued metrics available in literature offering a comprehensive evaluation methodology for XAI practitioners. System Causability Scale~\cite{holzinger2020measuring} is the other most prominent work in XAI evaluation. \textcolor{black}{Notably, all the above scales are designed to be application-domain agnostic, with each item adaptable to reference the specific AI system under evaluation.} We discuss each scale briefly below.   

\begin{description}
\item[Hoffman's Goodness Checklist] is utilised to objectively evaluate explanations with an independent XAI expert to improve the ``goodness''. It consists of 7 items answered by either selecting 'yes' or 'no'. It was developed by referring to literature that proposes ``goodness'' properties of explanations. 

\item [Hoffman Satisfaction Scale] was designed using psychometric theory to evaluate the subjective ``goodness'' of explanations with target users. It consists of 8 items responded in a 5-step Likert Scale. It is viewed as the user-centred variant of the Goodness Checklist with many shared items. This scale has been evaluated for content validity with XAI experts as well as construct and discriminant validity in pilot studies.  

\item [Hoffman Curiosity Checklist] is designed to elicit stakeholder explanation needs, i.e. which aspects of the system pique their curiosity. This metric consists of one question \textit{Why have you asked for an explanation? Check all that apply.} and the responses inform the design and implementation of the XAI system. 

\item [Hoffman Trust Scale] measures the development of trust when exploring a system's explainability. The authors derived this trust scale by considering the overlaps and cross-use of scales from trust scales in literature for measuring trust in autonomous systems~(not in the presence of explainability, e.g. trust between human and a robot)~\cite{jian2000foundations,adams2003trust,schaefer2013perception,cahour2009does}.

\item [System Causability Scale] measures the effectiveness,  efficiency and satisfaction of the explainability process in systems involving multi-shot explanations~\cite{holzinger2020measuring}. Derived from the widely-used System Usability Scale~\cite{brooke1996sus}, this scale comprises 10 items rated on a 5-step Likert scale. Notably, it includes items that measure stakeholder engagement, addressing a gap in previous scales designed for one-shot explainability settings. However, the validation of the scale is limited to one small-scale pilot study in the medical domain. 
\end{description}

\subsubsection{Other Dimensions}
Many other publications emphasised the need for user-centred XAI evaluations and explored evaluation dimensions. Two other dimensions considered by Hoffman et al.~\cite{hoffman2023measures} are mental model and performance concerning task completion. They recommended eliciting the mental model of stakeholders in think-aloud problem-solving and question-answering sessions. Performance is measured by observing the change in productivity and change in system usage. The evaluation of these dimensions requires metrics beyond questionnaire-based techniques. Another domain-agnostic survey finds many overlaps with Hoffman et al., defining 4 user-centred evaluation dimensions: mental model, usefulness and satisfaction, trust and reliance and human-task performance~\cite{mohseni2021multidisciplinary}.  
Zhou et al.,~\cite{zhou2021evaluating} summarise previous literature, emphasising three subjective dimensions - trust, confidence and preference that overlap with dimensions identified in~\cite{hoffman2023measures}. Conversely to Hoffman et al., they consider task completion to be an objective dimension in user-centred XAI evaluation. 

Carvalho et al., delineate characteristics of a human-friendly explanation in the medical domain, including some subjective or user-centred properties such as comprehensibility, novelty, and consistency with stakeholders' prior beliefs~\cite{carvalho2019machine}. Notably, consistency with stakeholders' prior beliefs aligns with the mental model from Hoffman et al.~\cite{hoffman2023measures}, while novelty can influence stakeholder engagement~\cite{holzinger2020measuring}. 
Nauta and Seifert~\cite{nauta2023co} recognise 12 properties of explanation quality for image classification applications. They identify three user-centred properties: context - how relevant the explanation is to the user; coherence - how accordant the explanation is with prior knowledge and beliefs; and controllability - how interactive and controllable the explanation is. In comparison to other literature, controllability aligns with engagement~\cite{holzinger2020measuring} and coherence aligns with the mental model~\cite{hoffman2023measures,carvalho2019machine}. Context can be associated with several properties such as curiosity, satisfaction, and preference~\cite{hoffman2023measures,zhou2021evaluating}.

These findings highlighted that there are many overlaps between evaluation dimensions identified in recent years. 
However, we highlight two main gaps in this current work: 1) there is no consensus in previous literature regarding the applicable metrics to measure these evaluation dimensions; and 2) the majority of the existing dimensions and metrics focus on evaluating individual explanations~\textcolor{black}{(i.e. single-shot), not multi-shot} XAI experiences.

\subsection{Initial item bank}

Informed item sampling from literature resulted in 33 items, including 7 from the Goodness Checklist~\cite{hoffman2023measures}; 8 from the Satisfaction Scale~\cite{hoffman2023measures}; 8 from the Trust Scale~\cite{hoffman2023measures}; and 10 from the System Causability Scale~\cite{holzinger2020measuring}. 
\textcolor{black}{As evidenced in the literature review, while these items covered constructs related to user-centred evaluation of single-shot explanations, they lacked explicit representativeness of multi-shot explanations. Recognising that \textit{interaction} is crucial to creating effective explanation experiences, the research team generated seven additional items to capture this construct. These items were designed to capture stakeholder perspectives on the interactive experience, a dimension explicitly not covered in previous XAI evaluation literature. This expanded the item pool to a total of 40 items.}

\textcolor{black}{The item pool was then standardised to a symmetric 5-point Likert Scale, ranging from ``I Strongly Agree'' to ``I Strongly Disagree''. The existing user-centred XAI scales identified in the literature review commonly adopt the 5-point format, offering a balance between granularity and minimising ambiguity between response options. Broader research in psychology has also shown via mathematical modelling that using a scale with more than five points offers no significant improvement in internal consistency reliability measured by Cronbach's alpha~\cite{lissitz1975effect}.}

\textcolor{black}{Finally, following similar approaches from recent literature in scale development~\cite{hasan2023psychometric,votipka2020building}, the item pool then underwent a 2-stage rigorous review and revision process. Each stage of the revision process was led by two authors, who proposed revisions, and subsequently validated by the remaining authors, who either accepted, rejected, or further revised them.
In the first stage, lexically identical items were removed, and items with significant semantic overlap were consolidated, resulting in 32 items. 
During the second stage, expressions were made more concise, non-suggestive, and aligned with the intended construct (i.e., measuring XAI experiences rather than explanations).
Rephrased items were further validated by iterating wording until unanimous agreement between authors to ensure clarity for stakeholders with varying knowledge levels and to eliminate suggestive wording to minimise potential bias.
The resulting 32 items formed the initial XEQ Scale~(included in Supplementary Material). }

\subsubsection{Evaluation Dimensions}
We reviewed evaluation dimensions from previous literature and consolidated XEQ items into four evaluation dimensions representing XAI experience quality: learning, utility, fulfilment, and engagement.
These dimensions are relevant to capturing personalised experiences for a given stakeholder.
We define them as follows:
\begin{description}
\item [Learning:] the extent to which the experience develops knowledge or competence; 
\item [Utility:] the contribution of the experience towards task completion; 
\item [Fulfilment:] the degree to which the experience supports the achievement of XAI goals; and
\item [Engagement:] the quality of the interaction between the user and the XAI system.
\end{description}
\noindent In the next sections, we describe the development, refinement and validation of the XEQ Scale following Psychometric Theory~\cite{nunnally2007psychometric}.

\section{XEQ Scale Development}
\label{sec:o2}
\textcolor{black}{This section presents the details and results of the expert-led evaluation of the initial set of 32 items to develop the XEQ scale, following the Content Validity Ratio~(CVR) method~\cite{lawshe1975quantitative,gehlbach2011measure}.}

\subsection{Study instrument}

\textcolor{black}{The study instrument consisted of three components: 1) an introduction to terminology; 2) three XAI experience examples; and 3) a structured survey for data collection.
The first component introduced the participants to the terminology used throughout the study~(also included in Table~\ref{tbl:gloss}). This is intended to establish a consistent mental model between experts from diverse XAI backgrounds.
Next, participants explored three XAI experiences curated to capture different application domains, end-user explanation needs and interaction modalities. 
Lastly, participants responded to the survey with three sub-sections: a) rating the 32 items regarding their relevance for measuring XAI experience quality on a 5-step Likert scale ranging from \textit{Not Relevant at All} to \textit{Extremely Relevant}; b) rating the 32 items regarding their clarity on a similar Likert scale ranging from \textit{Not Clear at All} to \textit{Extremely Clear}; and c) rephrasing items that they found relevant but not clear. }

\subsubsection{\textcolor{black}{XAI experiences development and selection}}
\textcolor{black}{The development of sample XAI experiences was conducted over three stages. First, we established several desiderata for sample experiences: a) reflect diverse, real-world AI systems where explaining AI outcomes is critical; b) possess the level of complexity that is suited for multi-shot explanation; and c) capture varying levels of domain expertise and AI familiarity.}

\textcolor{black}{
The second stage builds on these desiderata, to identify relevant application domains for sample XAI experiences considering previous literature on XAI design and development, and feedback from industry and academic partners.
We selected radiograph fracture detection from the medical domain, welfare application approval auditing from the governance domain, and student support for course selection from academia.
Regulations governing the use of AI decision-support in medical imaging underscore the need for XAI, particularly through comprehensive explanation methodologies~\cite{jin2023guidelines,miro2022evaluating,muller2022explainability,lang2022artificial}. Similar regulations are being implemented for AI in civic and governmental decision-making~\cite{de2019and} to ensure that automated decisions are interpretable, contestable, auditable, and fair~\cite{de2022perils}. Finally, student support represents an area with significantly lower legal/ethical requirements for explainability, but with high user experience impact~\cite{brdnik2023assessing}.}

\textcolor{black}{
In the third and final stage, we designed each of the sample experiences. For each experience, the target user was identified to ensure a variety of AI knowledge and domain expertise levels. The interaction modality was selected to suit the target user's explanation and interaction needs. The sample interactions articulated different types of explanations addressing the target user's explanation needs.
The three sample experiences are summarised in Table~\ref{tbl:apps}. 
A sample dialogue with the interactive CourseAssist chatbot is shown in Figure~\ref{fig:exps} positive XAI experience. This dialogue, in its entirety, captures an XAI experience that we wish to evaluate. The experiences constructed for medical imaging and welfare application approval auditing are available in the Supplementary Material.}

\begin{table*}[ht]
\caption{\textcolor{black}{Sample XAI experiences; for each experience, we summarise the AI system, attributes of the target stakeholder and the explanations presented in their XAI experience. Knowledge levels are adapted from the Dreyfus 6-stage skill acquisition model~\cite{rousse2021revisiting}.}}
\label{tbl:apps}
\footnotesize
\renewcommand{\arraystretch}{1.15}
\centering
\begin{tabular}{llllll}
\hline
Identifier&Attributes&Name&Domain knowledge&AI familiarity&Interaction Modality\\
\hline
CourseAssist&\textbf{Stakeholder}&Student&Novice&Novice&Conversation\\
\cline{3-6}
& \textbf{AI System} & \multicolumn{4}{l}{A rule-based agent that recommends courses given a student's chosen career path}\\
& \textbf{Explanations} & \multicolumn{4}{l}{Distribution and statistics of the data used in the recommendations;}\\
&& \multicolumn{4}{l}{Factual explanations; Nearest-neighbour cases}\\
\hline
RadioAssist& \textbf{Stakeholder}&Trainee Radiologist&Proficient&Competent&Graphical User Interface \\
\cline{3-6}
& \textbf{AI System} & \multicolumn{4}{l}{A neural network black-box that predicts the presence of fracture given } \\
&& \multicolumn{4}{l}{a radiograph in PNG format}\\
&\textbf{Explanations} & \multicolumn{4}{l}{Distribution and statistics of the data used to develop the model; Feature attributions;} \\ 
&& \multicolumn{4}{l}{Nearest-neighbour cases; Performance metrics of the AI model}\\
\hline
AssistHub& \textbf{Stakeholder}&Auditor&Expert&Proficient&Graphical User Interface \\
\cline{3-6}
& \textbf{AI System} & \multicolumn{4}{l}{A decision tree algorithm that recommends the outcome given a welfare application} \\
& \textbf{Explanations} & \multicolumn{4}{l}{Distribution and statistics of the data used to create the AI system; }\\
&&\multicolumn{4}{l}{Prototypical approved and rejected cases; Outlier cases; Simplified decision tree} \\
\hline
\end{tabular}
\end{table*}

\subsection{Participant recruitment and selection}

\textcolor{black}{The CVR method recommends a small group of domain experts~(5-10) for quantifying the strength of psychometric scale items~\cite{lawshe1975quantitative}. We invited 38 XAI experts with diverse expertise within XAI, considering those who are actively publishing in the XAI domain since 2020. The study recruited a diverse group of experts, 3 from industry and 10 from academia. Their specific areas of expertise were identified as follows:
3 experts in implementing XAI in industrial engineering applications (Energy and Telecommunications); 
1 expert in XAI evaluation scale development; 
5 experts in example-based and counterfactual XAI methods; 
2 experts in human-centric XAI design and development; 
and 2 experts in the role of causality in XAI. 
They collectively represent the diverse aspects of crafting effective XAI experiences.}

\subsection{Data collection and analysis}
\textcolor{black}{The study was hosted on the Jisc Online Surveys platform for 3 weeks and collected responses from 13 XAI experts. Their responses were collated and analysed using the following metrics.}

\subsubsection*{Content Validity}
The Content Validity Index (CVI) assesses item validity based on responses to the relevance property. Lower scores indicate items that may need modification or removal.
Given scale $\mathbf{S}$ with $M$ items where $i$ indicates an item, $r^i_j$ denotes the response of participant $j$ to item $i$.
For analysis, each response~($r^i_j$) is modified as follows. 
\begin{equation*}
r^i_j = 
\begin{cases}
    1,& \text{if } r^i_j \in [\text{\footnotesize Extremely Relevant or  Somewhat Relevant}]\\
    0, & \text{otherwise}
\end{cases}
\end{equation*}
We calculate the following two forms of the Content Validity Index~(CVI) scores.
\begin{description}
\item [Item-Level CVI:] measures the validity of each item independently; the number of responses is $N$ and the expected score is $\geq 0.78$.
\[
I\text{-}CVI_i = \frac{\sum_{j=1}^N(r^i_j)}{N}
\]
\item [Scale-Level CVI:] measures the overall scale validity using a) Average method i.e. the mean Item-Level CVI score where the expected score is $\geq0.90$; and b) Universal Agreement method i.e. the percentage of items experts always found relevant with the expected value of $\geq0.80$.
\[
S\text{-}CVI(a) = \frac{\sum_{i=1}^M(I\text{-}CVI_i)}{M}
\]
\[
S\text{-}CVI(b) = \frac{\sum_{i=1}^M 1_{[I\text{-}CVI_i = 1]}}{M}
\]
Here, once the average of the I-CVIs is calculated for all items with S-CVI$(a)$, S-CVI$(b)$ counts the number of items with an I-CVI of 1 (indicating complete agreement among experts that the item is relevant) and divides this by the total number of items.
\end{description}


\subsection{Results}
We refer to columns Item and I-CVI from Table~\ref{tbl:results-ap} for the results of the Content Validity study.
We first removed items with low validity ($\text{I-CVI}_i\leq 0.75$) and thereafter S-CVI scores were used to establish the content validity of the resulting scale.
Here we marginally divert from the established baseline of $0.78$ for I-CVI to further investigate items with $0.75 \leq \text{I-CVI} \leq 0.78$ during the pilot study. The Likert responses to the clarity property and free text feedback influenced the re-wording of 7 items to improve clarity~(indicated by $\dagger$). The item selection and rephrasing were done based on the suggestions from the XAI experts and the consensus of the research team. 
The resulting scale comprised 18 items, which we refer to as the \textit{XEQ Scale}. 
In Table~\ref{tbl:results-ap}, items are ordered by their I-CVI scores. 

S-CVI$(a)$ and S-CVI$(b)$ of the scale were 0.8846 and 0.2222. While S-CVI$(a)$ is comparable to the baseline of $0.9$, S-CVI$(b)$ indicate universal agreement is not achieved. However, existing literature suggests that meeting one of the baseline criteria is sufficient to proceed to pilot studies.
Notably, the 14 items with I-CVI $\geq 0.78$ also only achieve average agreement~(S-CVI$(a)=0.9179$) and not universal agreement~(S-CVI$(b)=0.2667$).

Following the item selection, each item was assigned an evaluation dimension based on the consensus of the research team~(see Column ``Evaluation Dimension'' in Table~\ref{tbl:results-ap}). These will be used in further investigations using \textcolor{black}{confirmatory} factor analysis \textcolor{black}{in Section~\ref{sec:o3}} to establish the construct validity.

\section{XEQ Scale Refinement and Validation}
\label{sec:o3}
In this section, we present the pilot study conducted to refine the XEQ Scale for internal consistency and construct validity\textcolor{black}{,} discriminant validity \textcolor{black}{and test-retest reliability}.

\subsection{Study instrument}
\textcolor{black}{The pilot study instrument consisted of three components: 1) an introduction to the terminology; 2) the explanation experience; and 3) a structured survey for data collection. Similar to the CVR study, introducing the terminology is intended to establish a common understanding among participants. Next participants explored a sample XAI experience stepping into the role of the target user. Finally, participants evaluated the sample XAI experience by responding to the 18 items each on a 5-step Likert scale ranging from \textit{I strongly agree} to \textit{I strongly disagree}}.

\subsubsection{\textcolor{black}{XAI experiences development and selection}}
\textcolor{black}{The pilot study is large-scale~(100+ participants) and aims, among other goals, to establish discriminant validity. To achieve this, we identified the following practical considerations for designing sample experiences: a) ensuring access to a large pool of target users via survey recruitment platforms or within the research institute; b) curating a relatively negative experience counterpart for each sample for validating the scale's discriminant properties in a controlled setting; and c) ensuring access to the same sample over longitudinal follow-up to validate the scale’s test-retest reliability.}

\begin{figure*}
    \centering
    \includegraphics[width=.98\textwidth]{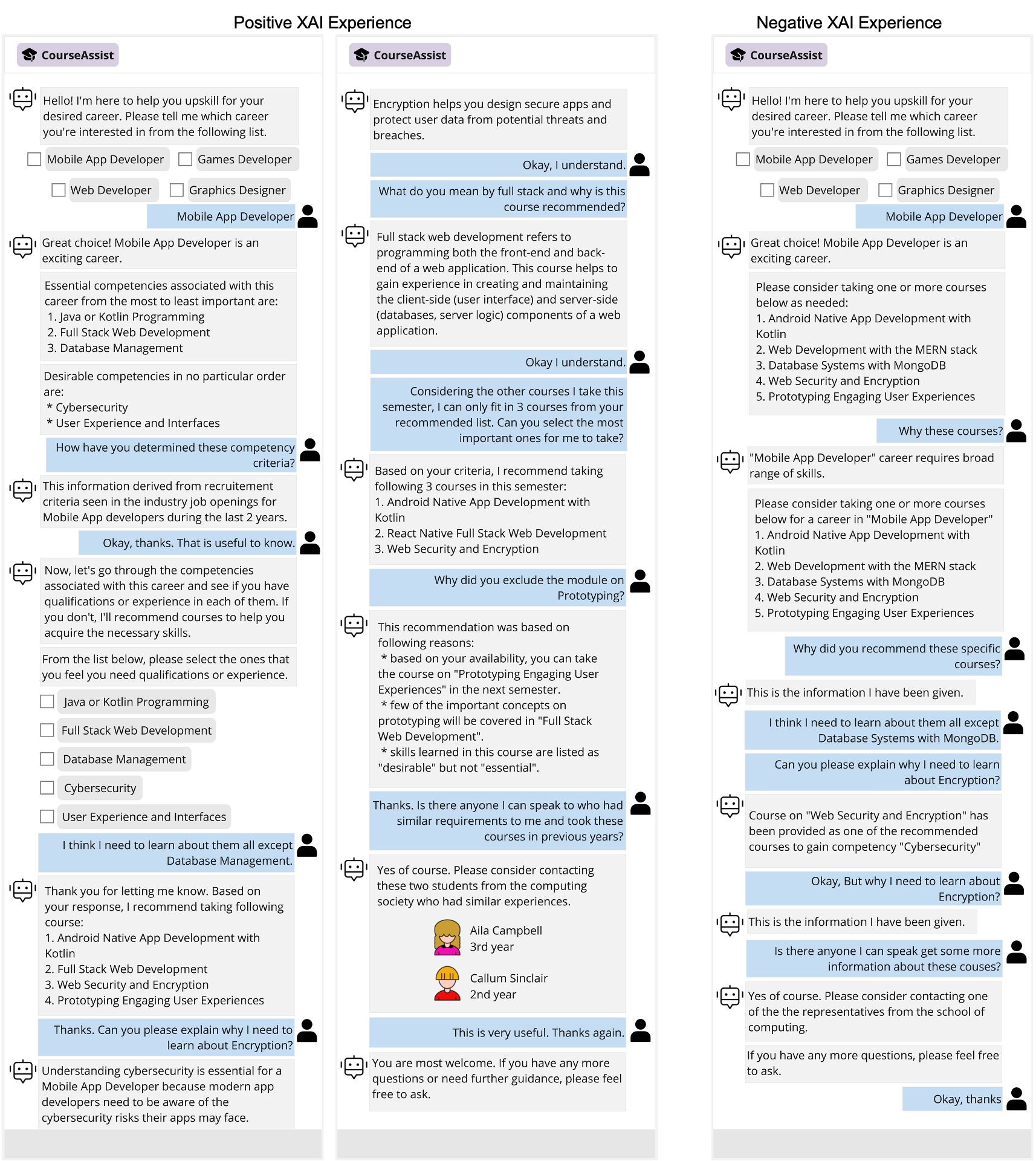}
    \caption{The static previews of the relatively positive~(left 2 columns) and negative~(right column) XAI Experiences with the CourseAssist Chatbot designed for the Pilot Study.}
    \label{fig:exps}
\end{figure*}

\textcolor{black}{Based on these considerations, we selected the CourseAssist and AssistHub systems for sample experiences from the point of view of a student and a welfare applicant respectively. Negative experiences, in contrast to positive ones, were deliberately designed to lack best practices recommended in recent human-centred XAI literature~\cite{liao2020questioning,suresh2021beyond,malandri2023convxai,nimmo2024user}. Specifically, they lacked: a) diversity in explanations within the experience; b) alignment between the target user's intent and the explanations provided; c) alignment between the user's knowledge level and the explanations; and d) mechanisms to encourage engagement with the XAI agent. 
The resulting four sample experiences are identified as CourseAssist-P, CourseAssist-N, AssistHub-P, AssistHub-N respectively where P and N indicate positive and negative experiences. Figure~\ref{fig:exps} presents the static views of CourseAssist-P and CourseAssist-N. AssistHub samples are included in Supplementary Material.}

\subsection{Participant recruitment and selection}
\textcolor{black}{The inclusion and exclusion criteria for participant recruitment to evaluate the CourseAssist and AssistHub experiences are summarized in Table~\ref{tbl:rct}. 
After getting through the inclusion criteria, each participant was randomly assigned to assess either the positive or negative sample experience in their respective application. After preliminary review, we excluded 5, 1 and 1 responses from groups CourseAssist-P, AssistHub-P and AssistHub-N who failed the following attention checks: 1) spend less than half of the allocated time; and/or 2) responded to the questionnaire in a pattern.
This yielded 68, 70, 50, and 50 responses for the CourseAssist-P, CourseAssist-N, AssistHub-P, and AssistHub-N cohorts.}

\begin{table*}[ht]
\caption{\textcolor{black}{Recruitment details for the pilot study; The only exclusion criterion was that participants of the CourseAssist study were not eligible for AssistHub, and vice versa.}}
\label{tbl:rct}
\footnotesize
\renewcommand{\arraystretch}{1.15}
\centering
\begin{tabular}{llp{7cm}lr}
\hline
Identifier&Stakeholder&Inclusion criteria&Source&Sample\\
\hline
CourseAssist&Student&Age -- between 18 and 30; Current education level -- Undergraduate degree; &Institute&68\\
&&Degree subjects -- Mathematics and statistics, Information and Communication &Prolific&70\\
&&Technologies, Natural sciences; Language of instruction -- English\\
\hline
AssistHub&Welfare applicant&Age -- above 30; Household size -- 3 or larger; &Prolific&100\\
&&Property ownership -- social housing or affordable-rented accommodation; \\
&&Employment status -- part-time, due to start a new job within the next month, \\
&&unemployed, or not in paid work (e.g. homemaker or retired)\\
\hline
\end{tabular}
\end{table*}

\subsection{Data collection and analysis}

\textcolor{black}{The study was hosted on the Jisc Online Surveys platform. Internal consistency, discriminant validity and construct validity were evaluated with both CourseAssist and AssistHub experiences with 238 participants. Among the 68 participants from the research institute who evaluated CourseAssist experiences, 35 took part in the test-retest validity study. AssistHub experiences were excluded from the test-retest validity study due to the practical challenges of following up and re-identifying participants for retesting on the Prolific platform. The duration between the test and retest was 4 weeks}. 

For analysis, we introduce the following notations. Given $r^i_j$ is the participant $j$'s response to item $i$, the participant's total is $r_j$ and the item total is $r^i$.
We transform 5-step Likert responses to numbers as follows: Strongly Disagree-1, Somewhat Disagree-2, Neutral-3, Somewhat Agree-4, and Strongly Agree-5. 
Accordingly, for the 18-item XEQ Scale, $r_j \leq 90$ (5$\times$18).

\subsubsection{Internal Consistency}
Internal consistency refers to the degree of inter-relatedness among items within a scale. We employ the following metrics from psychometric theory to assess the XEQ Scale items.

\begin{description}
\item [Item-Total Correlation] calculates the Pearson correlation coefficient between the item score and the total score the expected value per item is $\geq 0.50$. The Item-Total Correlation of item $i$, $iT$ is calculated as follows.
\[ iT = \frac{\sum_{j=1}^N{(r^i_j - \bar{r^i})(r_j - \bar{r})}}{\sqrt{\sum_{j=1}^N{(r^i_j - \bar{r^i})^2} \sum_{j=1}^N{(r_j - \bar{r})^2}}} \]
Here $\bar{r^i}$ is the average response score for the $i$th item, 
and $\bar{r}$ is the overall response average. 
\item [Inter-Item Correlation] is a measure of the correlation between different items within a scale and values between 0.2 and 0.8 are considered expected since $\geq 0.80$ indicate redundancy and $\leq 0.20$ indicate poor item homogeneity. 
The calculation is similar to the previous but is between two items.

\item [Cronbach's alpha] measures the extent to which all items in a scale are measuring the same underlying construct~\cite{cronbach1949essentials}. High internal consistency is indicated by $\alpha \geq 0.7$. If $s^i$ is the standard deviation of responses to item $i$, and $s$ is the standard deviation of response totals, $\alpha$ is calculated as follows.
\[
\alpha = \frac{M}{M-1}\left(1-\frac{\sum_{i=1}^M (s^i)^2}{s^2}\right)
\]
As such this helps to quantify how much of the total variance is due to the shared variance among items, which reflects their consistency in measuring the same underlying construct. 
\end{description}

\subsubsection{Discriminant Validity}
Discriminant validity measures the ability of the scale to discern between positive and negative experiences and we used the following two methods. 

\begin{description}

\item [Discriminant Analysis] treats the pilot study responses as a labelled dataset to train a classification model with a linear decision boundary. The items are considered as features and the group~(A or B) is considered as the label. A holdout set then evaluates the model's ability to distinguish between groups A and B. 

\item [Parametric Statistical Test] uses a mixed-model ANOVA test to measure if there is a statistically significant difference between the two groups A and B~(agnostic of the domain).  Our null hypothesis is ``no significant difference is observed in the mean participant total between groups A and B``. Our sample sizes meet the requirements for a parametric test determined by an a priori power analysis using G*Power~\cite{faul2007g}. 

\end{description}

\subsubsection{Construct Validity}
Construct validity evaluates the degree to which the scale assesses the characteristic of interest, i.e. factors~\cite{kazdin2021research}. Via factor analysis we aim to uncover underlying factors~(i.e. dimensions) and validate them. Two forms of factor analysis applied are:

\begin{description}

\item [Exploratory Factor Analysis (EFA)] finds the number of underlying factors in the scale by assessing the variance explained through the Principal Component Analysis~(PCA) coefficients~(i.e. eigenvalues).

\item [Confirmatory Factor Analysis (CFA)] tests a pre-defined factor model, with factor loadings expected to be >0.5 to indicate strong support. \textcolor{black}{For the XEQ scale, the evaluation dimensions assigned in Table~\ref{tbl:results-ap} are validated as a 4-factor model}.

\end{description}

\subsubsection{
\textcolor{black}{Test-retest Reliability}}
\textcolor{black}{The consistency and stability of the XEQ scale over time are measured with the test-retest reliability. We provide quantitative evidence by applying the following metrics.}  

\begin{description}

\item [\textcolor{black}{Pearson Correlation Coefficient} ] 
\textcolor{black}{measures the correlation between scale scores at test and retest. 
Let $r_j$ and $r_j'$ represent participant 
$j$'s scale scores at the first instance (test) and the second instance (retest), respectively, with $\bar{r}$ and $\bar{r'} $ being the mean scale scores for all participants at each time point. A $\rho > 0.7$ indicates strong test-retest reliability, suggesting high consistency between the two measurement instances. 
}
\[ \rho = \frac{\sum_{j=1}^N{(r_j - \bar{r})(r_j' - \bar{r'})}}{\sqrt{\sum_{j=1}^N{(r_j - \bar{r})^2} \sum_{j=1}^N{(r_j' - \bar{r'})^2}}} \]

\item [\textcolor{black}{Intra-class Correlation Coefficient (ICC)}] \textcolor{black}{assesses the consistency of scale scores from the same group of participants between test and retest, using a two-way mixed-effects model~(i.e. ICC3).
Given that each participant $j$ has a mean score of $\bar{r_j}$ between test and retest, and the overall mean across all participants is $\overline{\overline{r}}$ the ICC is computed as:}
\[ 
ICC = \frac{ms - ms_e}{ms + (k-1)\times ms_e}
\]
, \textcolor{black}{where} $ms$ \textcolor{black}{is the mean square between participants, }$ms=\sum_{j=1}^N{(\bar{r_j} - \overline{\overline{r}})^2}/(N-1)$ and $ms_e$ \textcolor{black}{is the mean square error }
$ms_e=\sum_{j=1}^N{[(r_j - \bar{r_j})^2 + (r_j' - \bar{r_j})^2]}/[N \times (k-1)]$. \textcolor{black}{An ICC value above 0.75 indicates good reliability, and above 0.90 reflects excellent reliability, indicating strong consistency in scale scores between test and retest.}

\end{description}

\begin{table*}[!t]
\centering
\caption{\textcolor{black}{Results: 18 items were selected from the item bank based on the Item-Level Content Validity Index(I-CVI> 0.75) and were assigned an evaluation dimension. From the pilot studies, we find Item-Total Correlation (iT)>0.5 for all 18 items. One-factor loadings all over 0.5 confirm the overarching single factor within the XEQ Scale. Confirmatory (C.) Factor Analysis shows significant loading (>0.5) from each item for their respective evaluation dimension.}}
\label{tbl:results-ap}
\footnotesize
\renewcommand{\arraystretch}{1.15}
\centering
\begin{tabular}{lp{6cm}rrrrrr}
\hline
\#&Item & I-CVI&$iT$&One-Factor&Evaluation&\textcolor{black}{C. Factor}\\
&&&&Loading&Dimension&\textcolor{black}{Loading}\\
\hline
1&The explanations received throughout the experience were consistent$^\dagger$.&1.0000&0.6274&0.6076&Engagement&\textcolor{black}{0.7988}\\
2&The experience helped me understand the reliability of the AI system.&1.0000&0.6416&0.6300&Learning&\textcolor{black}{0.9409}\\
3&I am confident about using the AI system.&1.0000&0.7960&0.7790&Utility&\textcolor{black}{1.0096}\\
4&The information presented during the experience was clear.&1.0000&0.7666&0.7605&Learning&\textcolor{black}{1.1634}\\
5&The experience was consistent with my expectations$^\dagger$.&0.9231&0.7959&0.7831&Fulfilment&\textcolor{black}{1.0822}\\
6&The presentation of the experience was appropriate for my requirements$^\dagger$. &0.9231&0.8192&0.8083&Fulfilment&\textcolor{black}{0.9300}\\
7&The experience has improved my understanding of how the AI system works.&0.9231&0.6169&0.5859&Learning&\textcolor{black}{0.8280}\\
8&The experience helped me build trust in the AI system.&0.9231&0.7160&0.7018&Learning&\textcolor{black}{0.9787}\\
9&The experience helped me make more informed decisions.&0.9231&0.7460&0.7279&Utility&\textcolor{black}{0.7909}\\
10&I received the explanations in a timely and efficient manner.&0.8462&0.7015&0.6841&Engagement&\textcolor{black}{0.8592}\\
11&The information presented was personalised to the requirements of my role$^\dagger$.&0.8462&0.7057&0.6801&Utility&\textcolor{black}{0.7938}\\
12&The information presented was understandable within the requirements of my role$^\dagger$.&0.8462&0.7876&0.7803&Utility&\textcolor{black}{1.1452}\\
13&The information presented showed me that the AI system performs well$^\dagger$.&0.8462&0.8112&0.8016&Fulfilment&\textcolor{black}{0.9071}\\
14&The experience helped to complete the intended task using the AI system.&0.8462& 0.8299&0.8241&Utility&\textcolor{black}{1.0787}\\
\hline
15&The experience progressed sensibly$^\dagger$. &0.7692&0.8004&0.7912&Engagement&\textcolor{black}{1.1588}\\
16&The experience was satisfying.&0.7692&0.7673&0.7529&Fulfilment&\textcolor{black}{1.0309}\\
17&The information presented during the experience was sufficiently detailed.&0.7692&0.8168&0.8035&Utility&\textcolor{black}{1.0299}\\
18&The experience provided answers to all of my explanation needs.&0.7692&0.8472&0.8444&Fulfilment&\textcolor{black}{0.9041}\\
\hline
\end{tabular}
\end{table*}

\subsection{Results}

\subsubsection{Internal Consistency} 
Table~\ref{tbl:results-ap} column $iT$ reports the Item-Total Correlation. All items met the baseline criteria of $iT \geq 0.5$ and baseline criteria for Inter-Item correlation. Cronbach's alpha is $0.9562$ which also indicates strong internal consistency. 

\subsubsection{Discriminant Validity}
We performed discriminant analysis over 100 trials where at each trial a different train-test split of the responses was used. Each trial used a stratified split, with 70\% of the responses for training and 30\% for testing. Over the 100 trials, we observed accuracy of $0.63\pm0.05$ and a macro F1-score of $0.63\pm0.05$ which is significantly over the baseline accuracy of $0.50$ for a binary classification task. 
Mixed-model ANOVA test showed a statistically significant difference between groups P and N with a p-value of $1.63e-12$ where the mean participant total for groups P and N were $70.96\pm0.47$ and $57.97\pm1.84$. Also, it revealed a substantial variability within groups indicated by the group variance of 104.86, which we account for including responses from two application domains. Furthermore, Cohen's d was 1.7639 which indicates a large effect size confirming a significant difference between groups A and B. A standard t-test also obtained a p-value of $1.13e-09$ further verifying the statistically significant difference. 
Based on this evidence we reject the null hypothesis and confirm the discriminant validity of the scale.

\subsubsection{Construct Validity}
\begin{figure}[!t]
    \centering
    \includegraphics[width=.6\textwidth]{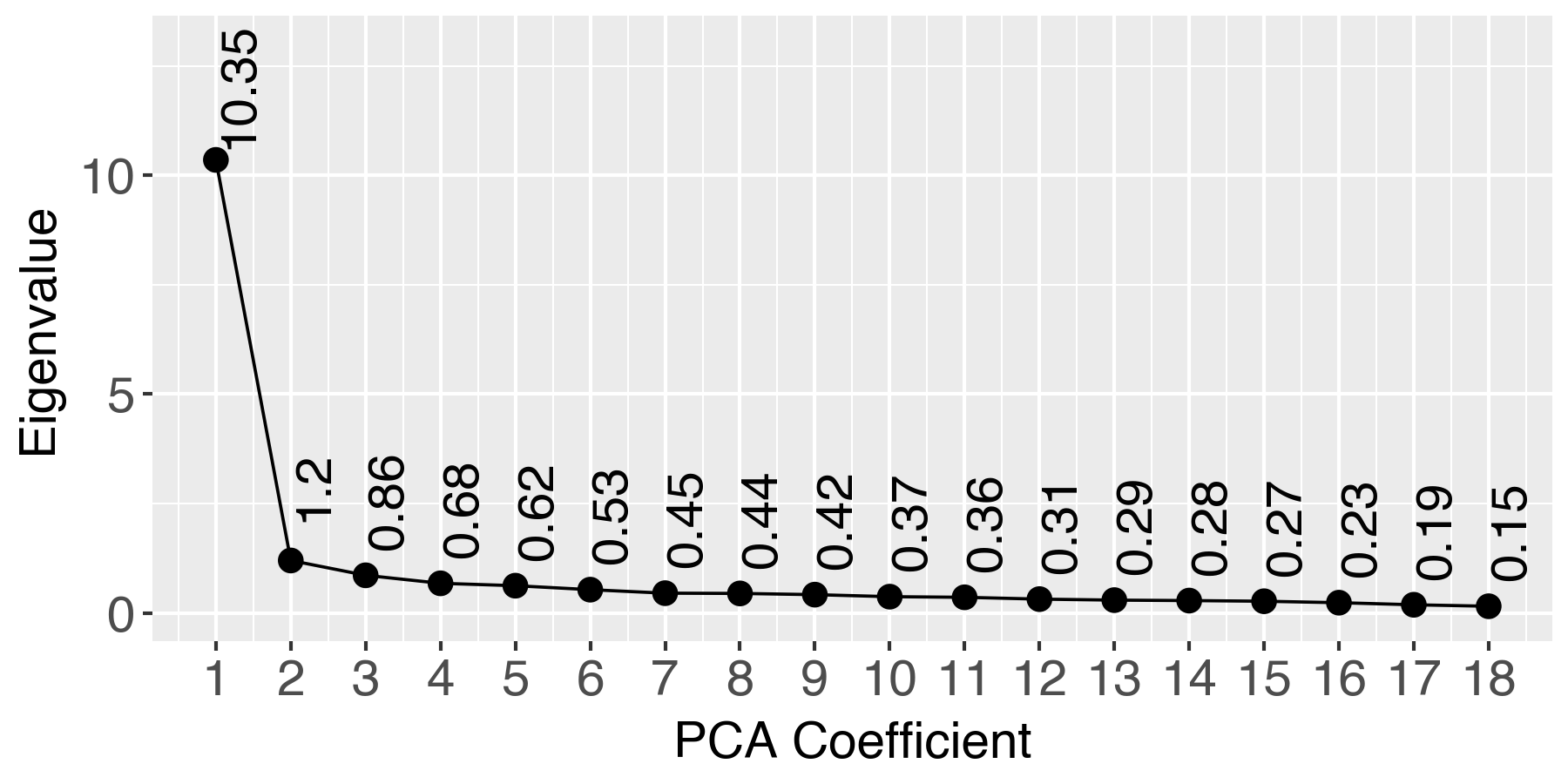}
    \caption{\textcolor{black}{Scree plot of eigenvalues derived from the Principal Component Analysis (PCA) performed for the exploratory factor analysis (EFA); a sharp decline in eigenvalues indicates the presence of a single overarching factor.}}
    \label{fig:efa}
\end{figure}

We first explore the number underlying factors in the XEQ Scale using EFA. Figure~\ref{fig:efa} presents the eigenvalues for PCA coefficients derived from scale responses which shows a sharp drop and plateau of eigenvalues for coefficient 2 onwards. This signifies a single overarching factor evident throughout the scale which we refer to as ``XAI Experience Quality''. We further confirm this single factor via one-factor loadings~(i.e. CFA with a single factor model) given in Table~\ref{tbl:results-ap} where all item supports are > 0.5. 

\textcolor{black}{An additional CFA was conducted to confirm that evaluation dimensions had substantial factor loadings from their respective items. The rightmost column in Table~\ref{tbl:results-ap} presents the CFA factor loadings. Each item meets the required factor loading of $\geq 0.5$ for their evaluation dimension. These observations reinforce that while there is an over-arching factor on ``XAI Experience Quality'', it is substantially underpinned by the four evaluation dimensions \textit{Learning}, \textit{Utility}, \textit{Fulfilment} and \textit{Engagement}.}

\subsubsection{\textcolor{black}{Test-retest Reliability}}\textcolor{black}{The test-retest reliability of the scale was assessed with a 4-week interval using the two CourseAssist experiences. Pearson correlation coefficient} $\rho = 0.7561$ \textcolor{black}{with a p-value of} $1.5e-07$, \textcolor{black}{indicated a strong linear relationship between the test and retest scores. Additionally,} $ICC = 0.8609$ \textcolor{black}{with a p-value of} $ 3.17e-08$, \textcolor{black}{falls within the range that indicates strong consistency. Both measures establish the test-retest reliability of the XEQ scale.}

\noindent This concludes our multi-faceted evaluation and refinement of the XEQ Scale based on pilot study results.

\section{Discussion}
\label{sec:discuss}

\subsection{Implications and limitations}
In psychometric theory, conducting a pilot study involves administering both the scale under development and existing scales to participants. The objective is to assess the correlation between the new scale and those found in existing literature, particularly in shared dimensions. However, our pilot studies did not incorporate this, since to the best of our knowledge there are no previous scales that measured the XAI experience quality or multi-shot XAI experiences. While the System Causability Scale~\cite{holzinger2020measuring} is the closest match in the literature, it was not applicable as it featured in the initial items bank.
Also, the current pilot studies had limited variability in application domains. To address this limitation, we are currently planning pilot studies with two medical applications: 1) fracture prediction in Radiographs; and 2) liver disease prediction from CT scans. In the future, we will further validate and refine the scale as necessary. 

\subsection{XEQ scale in practice}
\textcolor{black}{We propose the XEQ Scale as a tool to support the development and evaluation of interactive user-centric XAI systems. We anticipate calculating 3 aggregate metrics based upon responses to the 18 items in the scale:}
\begin{description}
\item [Stakeholder XEQ Score] quantifies individual experiences and is calculated as the mean of stakeholder's responses to all items.
\item [Factor Score] quantifies the quality along each evaluation dimension and is calculated as the mean of all responses to the respective subset of items. For XAI system designers, a lower factor score indicates the dimensions that need improvement.
\item [System XEQ Score] quantifies the XAI experience quality of the system as a whole and is calculated as the aggregate of Factor scores. The System XEQ Score helps the XAI system designers to iteratively develop a well-rounded system grounded in user experience. The designers can also choose to assign a higher weight to one or a subset of dimensions that they find important at any iteration. System XEQ Score can also be utilised by external parties~(e.g. regulators, government), either to evaluate or benchmark XAI systems.
\end{description}

\noindent \textcolor{black}{We have created the following user story to demonstrate the usefulness of the XEQ Scale for the iterative development and evaluation of an applied XAI system.}

\begin{quote}
\textcolor{black}{``\textit{Jane is the systems manager for a motor insurance company. The company have recently introduced an AI system for calculating an insurance quote using applicants' immediate driving history and previous claims. As part of the company's regulatory obligations, the AI system is supported by several explanation algorithms. However, the systems management team are having complaints from sales staff that they struggle to justify system outcomes using these explanations. Jane therefore decides to use the XEQ Scale to measure the XE for 2 stakeholder groups: customers and sales staff. They start by recruiting a representative user group for each role. Subsequently, each participant used the XAI system and scored their experience with the XEQ Scale.}''}
\end{quote}

\begin{quote}
\textcolor{black}{``
\textit{Jane collates the responses and calculates the stakeholder scores, factor scores and system scores for each role. When comparing system scores, Jane finds that sales staff consistently score the XAI system lower compared to customers. Looking at granular factor scores, they identify that the Utility factor score is particularly low for sales staff, specifically, item 14 ('The experience helped me to complete the intended task using the AI system') received consistently negative feedback. From these insights, Jane recognises that an additional explanation algorithm should be added to support sales staff in task completion. Jane updated the explanation algorithms and trials once again with a sample of sales staff, finding that the new system increased the Utility factor score and overall, led to an improved system XEQ score.}''}
\end{quote}

\noindent \textcolor{black}{From the above user story, it can be observed that the XEQ Scale can support the iterative development of XAI systems by facilitating the targeted update via user-centred evaluation.}

\subsection{XEQ Benchmark Development}
The next phase for the XEQ scale entails developing a benchmark for XAI experience quality. This process includes administering the XEQ scale to over 100 real-world AI systems that provide explainability to stakeholders and establishing a classification system. 
We are currently following the established benchmark maintenance policy of the User Experience Questionnaire~\cite{schrepp2017construction} where we develop and release an XEQ Analysis tool with the benchmark updated regularly.

We envision when the scale is administered to stakeholders of a new XAI system, the benchmark will categorise the new system based on the mean participant total in each evaluation dimension as follows - \textit{Excellent}: Within the top 10\% of XAI systems considered in the benchmark; \textit{Good}: Worse than the top 10\% and better than the lower 75\%; \textit{Above average}: Worse than the top 25\% and better than the lower 50\%; \textit{Below average}: Worse than the top 50\% and better than the lower 25\%; and \textit{Bad}: Within the 25\% worst XAI systems. Accordingly, the XEQ benchmark will enable XAI system owners to iteratively enhance the XAI experience offered to their stakeholders.

\section{Conclusions}
\label{sec:conc}
In this paper, we presented the XEQ scale. The XEQ scale provides a framework for the comprehensive evaluation of user-centred XAI experiences. It fills a novel gap in the evaluation of multi-shot explanations which is currently not adequately fulfilled by any other evaluation metric(s). Throughout this paper, we have described the development and validation of the scale following psychometric theory. We make this scale available as a public resource for evaluating the quality of XAI experiences. 
In future work, we plan to investigate the generalisability of the XEQ scale on additional domains, AI systems and stakeholder groups. Beyond this, we propose to establish a benchmark using the XEQ scale. Our goal is to facilitate the user-centred evaluation of XAI and support the emerging development of best practices in the explainability of autonomous decision-making.

\begin{acknowledgments}
The authors thank all participants in the CVR and pilot studies. 
iSee is an EU CHIST-ERA project that received funding for the UK from EPSRC under grant number EP/V061755/1, for Ireland from the Irish Research Council under grant number CHIST-ERA-2019-iSee, for France from the French National Research Agency under grant number ANR-21-CHR4-0004, and for Spain from the MCIN/AEI and European Union ``NextGenerationEU/PRTR'' under grant number PCI2020-120720-2.
\end{acknowledgments}

\bibliography{ref}

\appendix
\section*{Appendix}

\subsection*{A: Initial Items Bank}
Table~\ref{tbl:apinit} presents the initial item bank compiled from literature and reviewed by the research team. 
\begin{table*}[ht]
\centering
\caption{Initial Items Bank}
\label{tbl:apinit}
\footnotesize
\begin{tabular}{l}
\hline
Item\\
\hline
I like using the system for decision-making.\\
The information presented during the experience was clear.\\
The explanations received throughout the experience did not contain inconsistencies.\\
I could adjust the level of detail on demand.\\
The experience helped me make more informed decisions.\\
The experience helped me establish the reliability of the system.\\
I received the explanations in a timely and efficient manner.\\
The experience was satisfying.\\
The experience was suitable for the intended purpose of the system.\\
I was able to express all my explanation needs.\\
The experience revealed whether the system is fair.\\
The experience helped me complete my task using the system.\\
The experience was consistent with my expectations within the context of my role.\\
The presentation of the experience was appropriate.\\
The experience has improved my understanding of how the system works.\\
The experience helped me understand how to use the system.\\
The experience was understandable in the context of my role.\\
The experience helped me build trust in the system.\\
The experience was personalised to the context of my role.\\
I could request more detail on demand if needed.\\
I did not need external support to understand the explanations.\\
The experience was helpful to achieve my goals.\\
The experience progressed logically.\\
The experience was consistent with my understanding of the system.\\
The duration of the experience was appropriate within the context of my role.\\
The experience improved my engagement with the system.\\
The experience was personalised to my explanation needs.\\
Throughout the experience, all of my explanation needs were resolved.\\
The experience showed me how accurate the system is.\\
All parts of the experience were suitable and necessary.\\
The information presented during the experience was sufficiently detailed for my understanding of the domain.\\
I am confident in the system.\\
\hline
\end{tabular}
\end{table*}

\subsection*{B: Content Validity Study}

\begin{figure*}
    \centering
    \includegraphics[width=.8\textwidth]{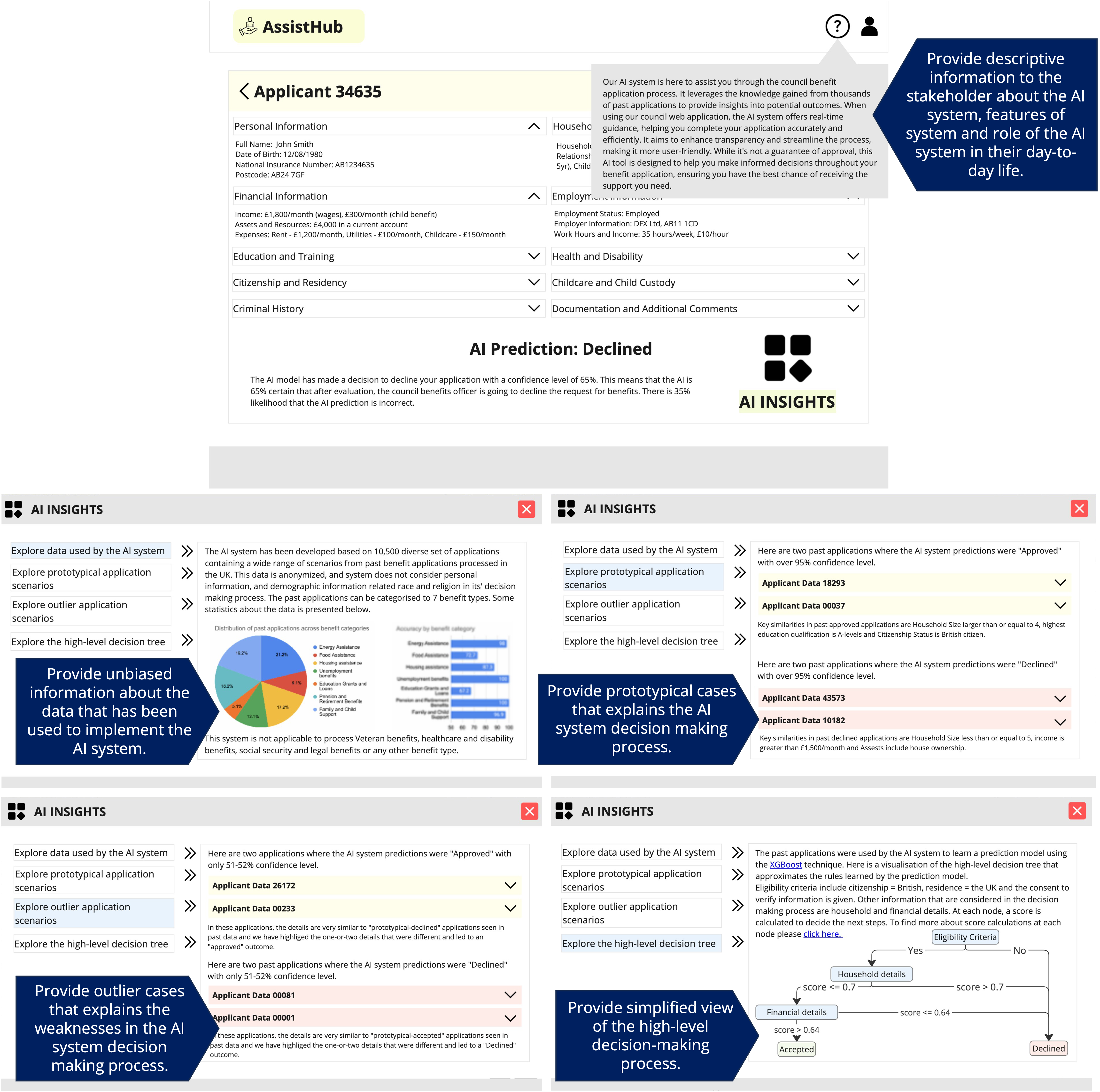}
    \caption{Positive XAI Experience of a Regulation Officer exploring the AssistHub AI platform; A pop-up is shown when clicked on the \textit{AI INSIGHTS} button, with four navigation pages providing different types of explanations. The help description pop-up appears when clicking on the question mark button.}
    \label{fig:welfare-cvr}
\end{figure*}

\begin{figure*}
    \centering
    \includegraphics[width=.6\textwidth]{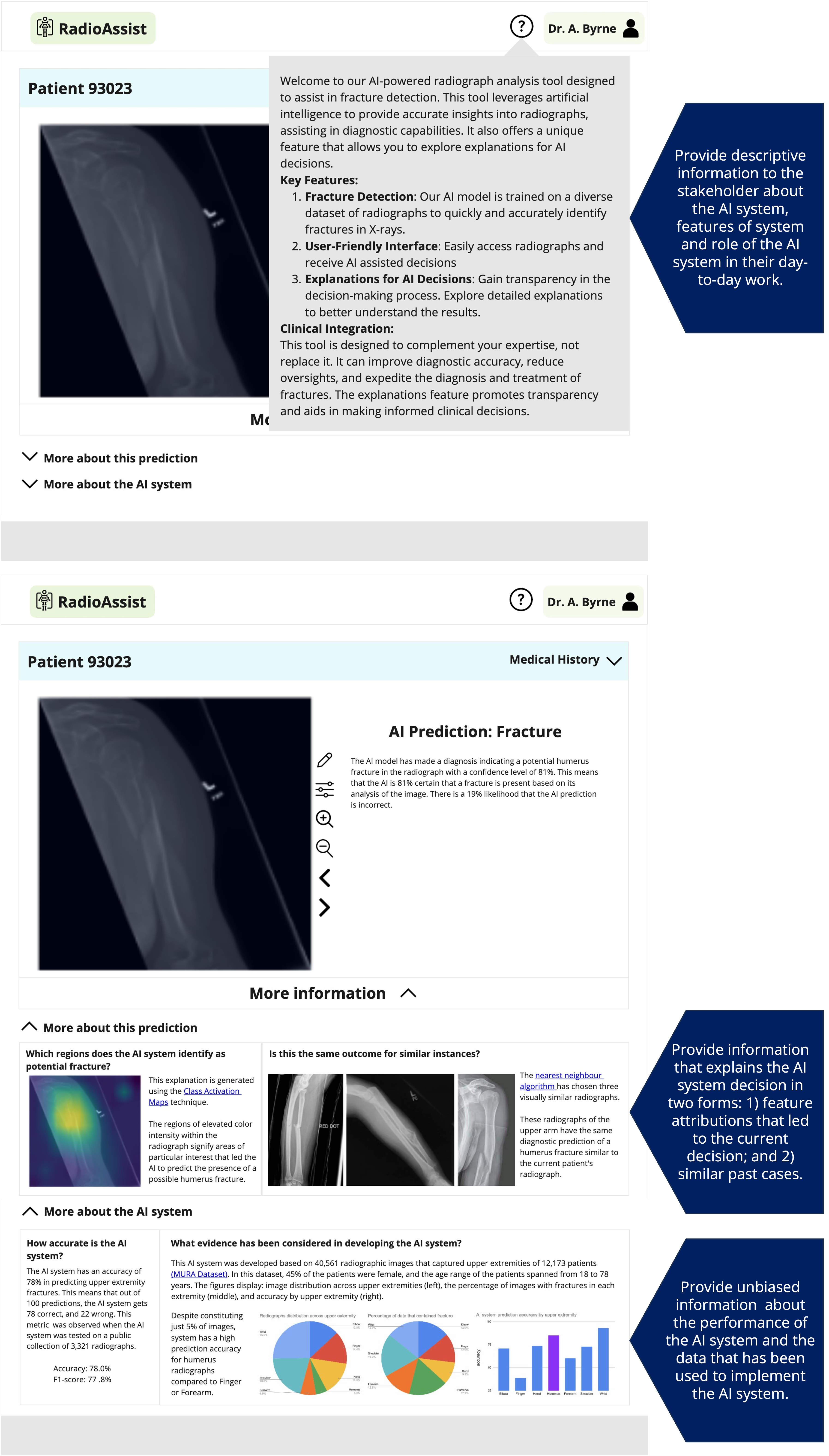}
    \caption{Positive XAI Experience of a Clinician using the RadioAssist AI platform; The clinician can click on the two minimised pages to expand and view explanations about the AI system and the decision. The question mark button shows the help description.}
    \label{fig:fracture-cvr}
\end{figure*}

\begin{figure*}
    \centering
    \includegraphics[width=.78\textwidth]{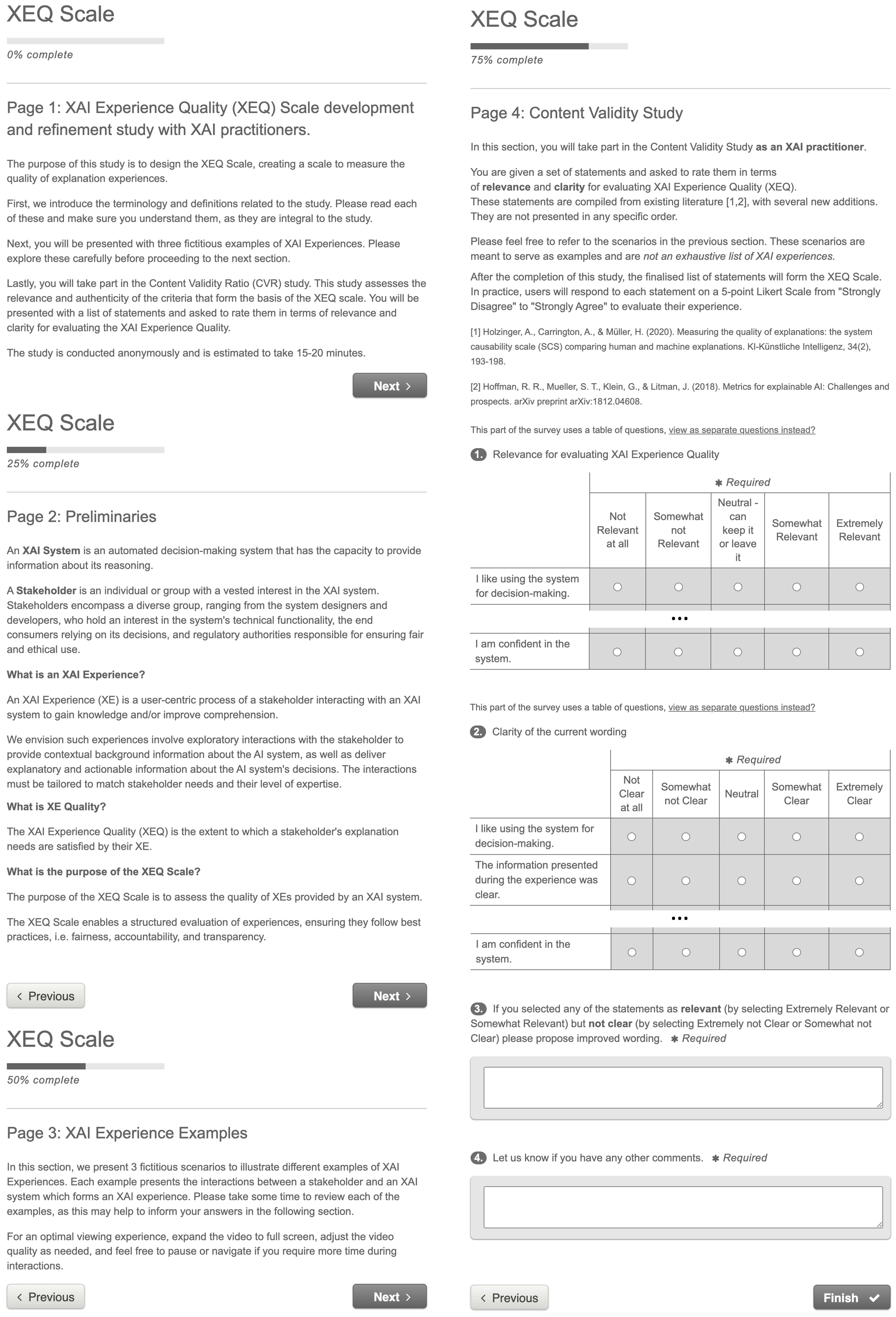}
    \caption{Study Preview; Examples removed from Page 3 and list of items shortened in Page 4. }
    \label{fig:cvr}
\end{figure*}

This study aimed to establish the content validity of the XEQ scale with XAI experts. Since XAI Experiences are a novel concept, we included three example XAI experiences that capture a variety of stakeholder types and application domains. 
In addition to the CourseAssist chatbot example included in the paper, they were presented with the following experiences in video format.
\begin{itemize}
\item The AssistHub AI platform is a website for processing welfare applications and is used by a local council to accelerate the application process. An auditor is exploring the website and its XAI features to understand the fairness and bias of the AI system being used in the decision-making process. A non-interactive preview of the experience is presented in Figure~\ref{fig:welfare-cvr}. 
\item RadioAssist AI platform is a desktop application, used by the local hospital to support clinicians in their clinical decision-making processes. An AI system predicts the presence of fracture in Radiographs and explains its decisions to the clinicians. A non-interactive preview of the experience is presented in Figure~\ref{fig:fracture-cvr}.  
\end{itemize}
Both Figures~\ref{fig:fracture-cvr} and~\ref{fig:welfare-cvr} are annotated with notes that describe the XAI features that were available to the stakeholder. 
Finally, Figure~\ref{fig:cvr} presents a preview of the Study page. 


\subsection*{C: Pilot Study}
\begin{figure*}
    \centering
    \includegraphics[width=.8\textwidth]{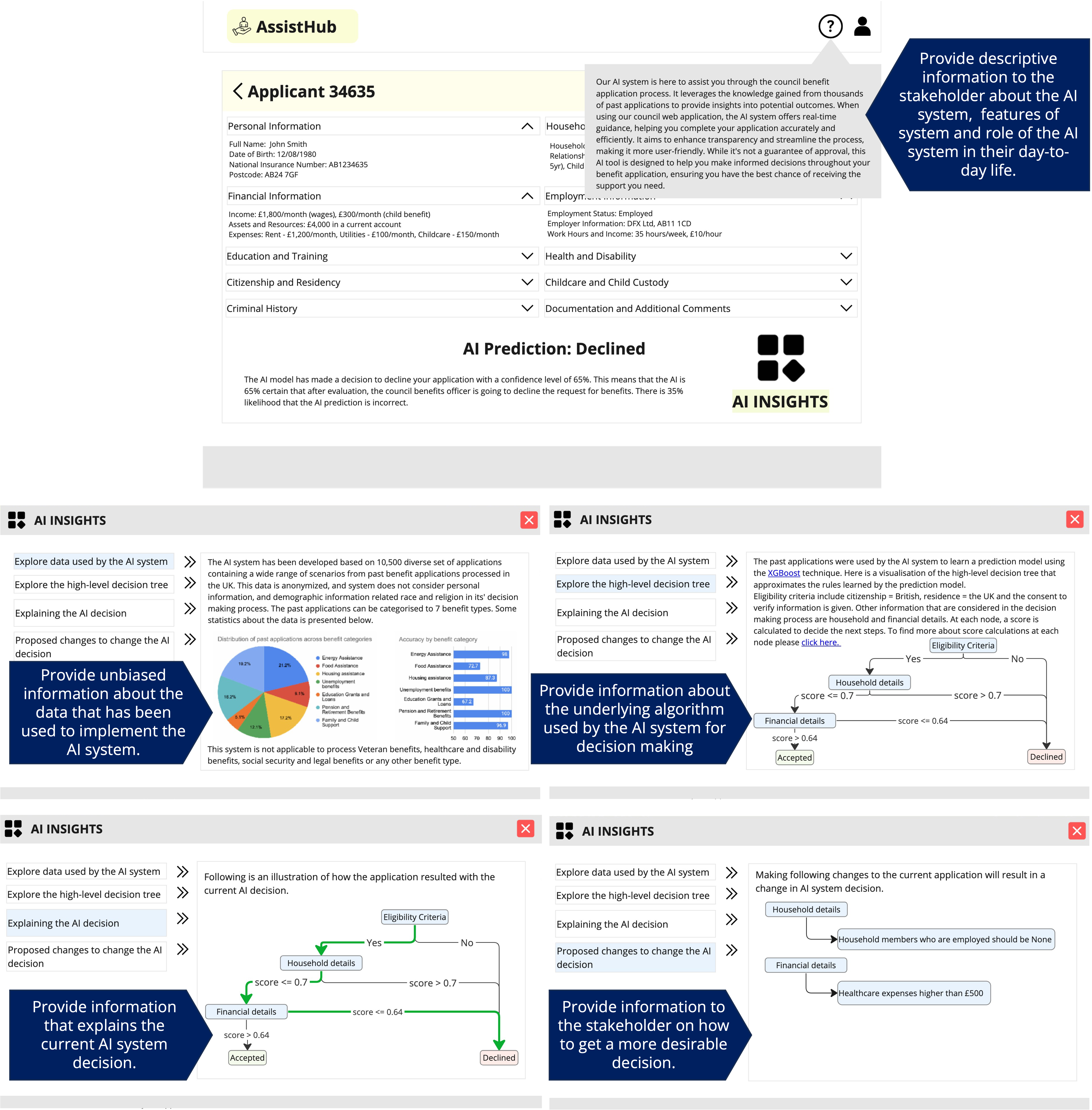}
    \caption{Relatively positive XAI Experience of a welfare applicant using the AssistHub AI platform}
    \label{fig:welfare-good}
\end{figure*}

\begin{figure*}
    \centering
    \includegraphics[width=.8\textwidth]{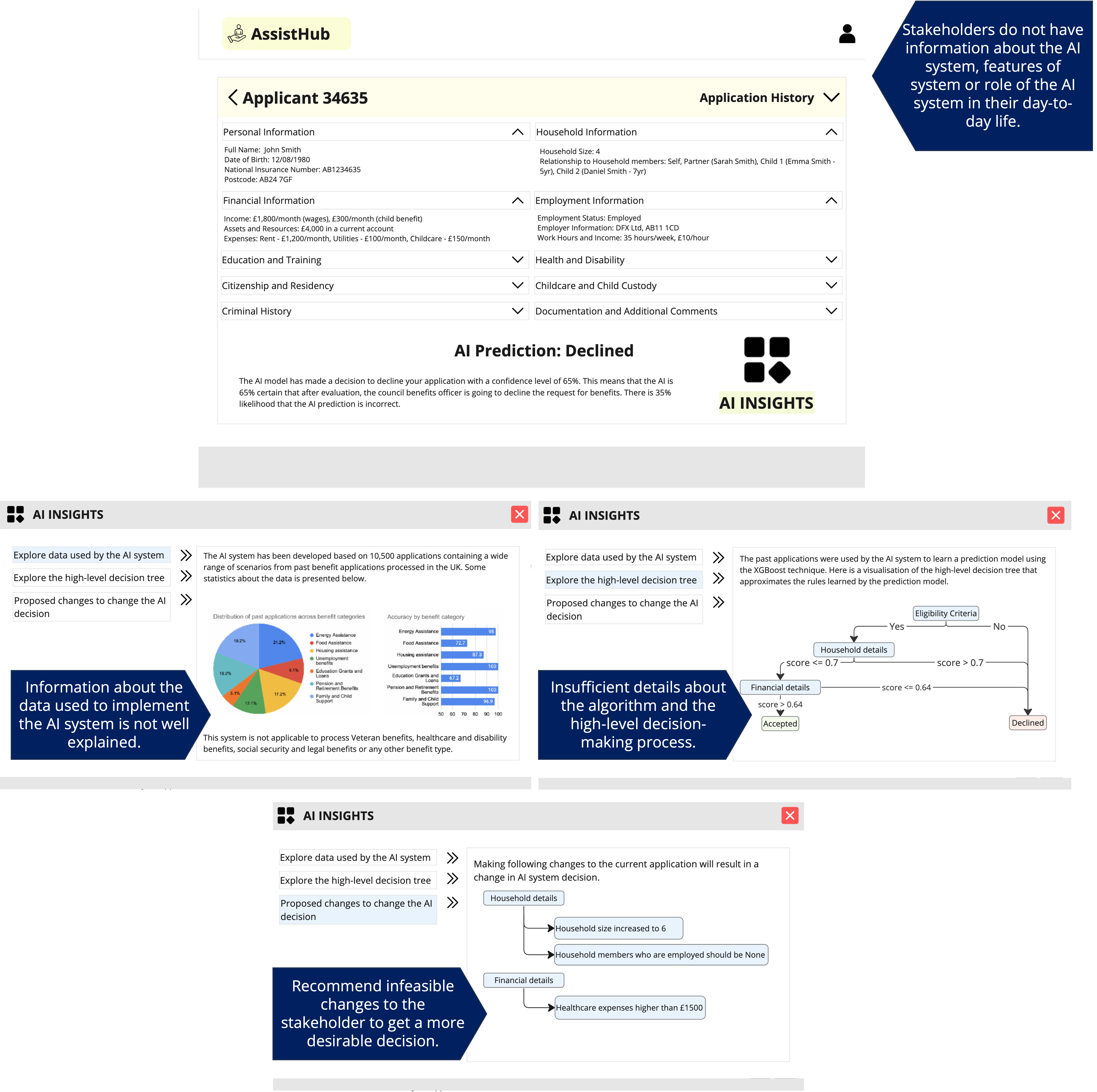}
    \caption{Relatively negative XAI Experience of a welfare applicant using the AssistHub AI platform}
    \label{fig:welfare-bad}
\end{figure*}

\begin{figure*}
    \centering
    \includegraphics[width=.78\textwidth]{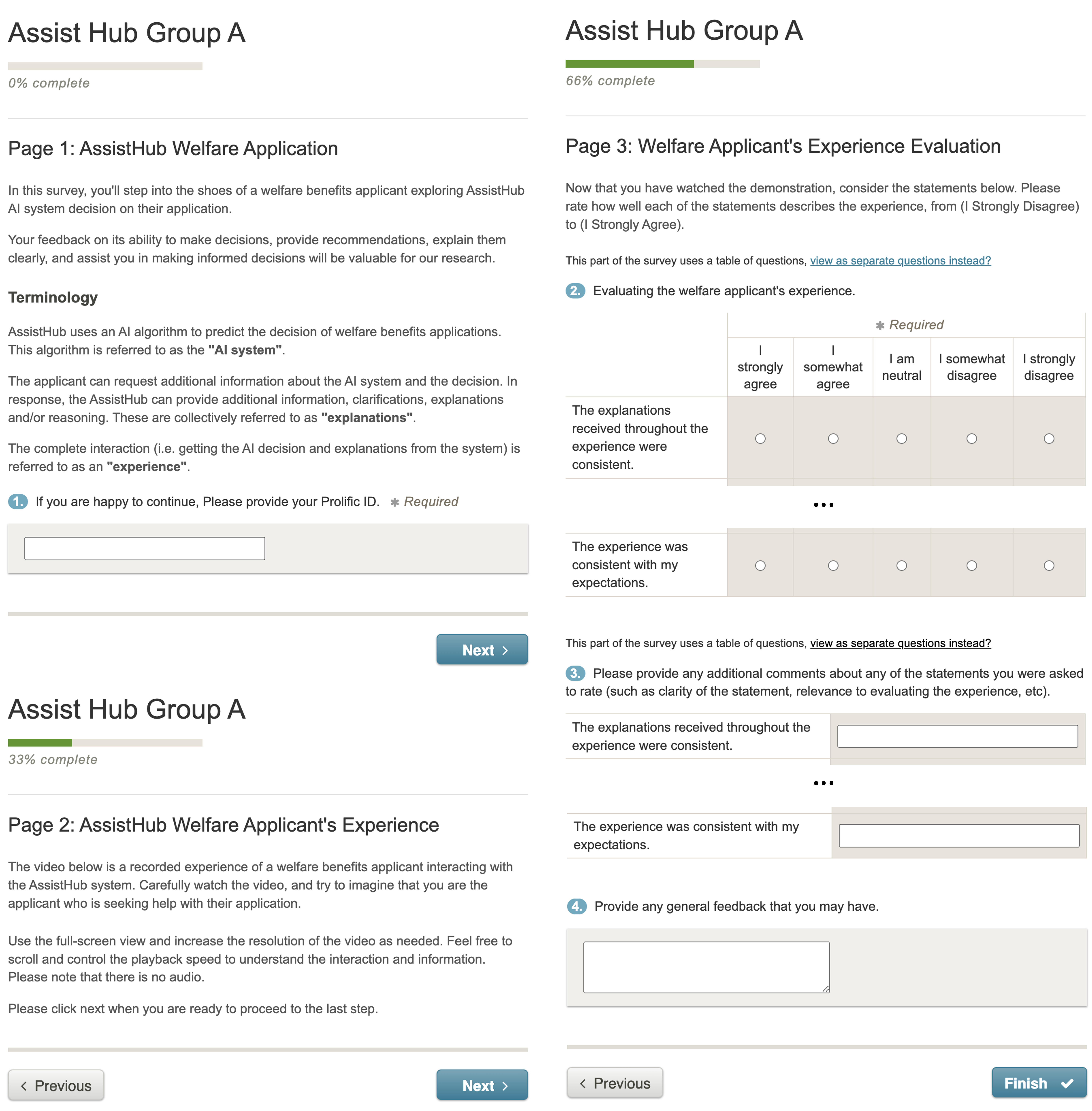}
    \caption{Pilot Study Preview; Example removed from Page 2 and list of items shortened in Page 3.}
    \label{fig:pilot}
\end{figure*}

A pilot study was conducted with 238 participants over two application domains where they evaluated either a positive or negative XAI experience. 
In addition to the CourseAssist chatbot examples provided in the paper, we included two XAI experiences of welfare applicants interacting with the AssistHub AI platform~(see Figures~\ref{fig:welfare-good} and~\ref{fig:welfare-bad}). Notes refer to how different aspects of the explanations can lead to a positive or negative XAI experience. Similar to the previous study, all XAI experiences were available to participants in video format.
Finally Figure~\ref{fig:pilot} presents a preview of the Pilot study where pages 1 and 2 were customised based on the application participants were assigned to. 

\end{document}